\documentclass[letterpaper, 10 pt, conference]{ieeeconf}  

\IEEEoverridecommandlockouts                              

\title{\LARGE \bf Improved Trajectory Reconstruction for Markerless Pose Estimation}

\usepackage{amsmath}
\usepackage{amsfonts}
\usepackage{booktabs}
\usepackage{xcolor}
\usepackage{multirow}
\usepackage[figuresright]{rotating}
\RequirePackage[colorlinks=true, allcolors=blue]{hyperref}

\usepackage[
    backend=biber,
    style=ieee,
    natbib=true,
    doi=false,
    url=false,
    isbn=false,
]{biblatex}

\AtEveryBibitem{%
   \clearfield{Title}%
   \clearfield{eprint}%
   \clearfield{note}%
}

\newbox{\myorcidaffilbox}
\sbox{\myorcidaffilbox}{\large\includegraphics[height=1.1ex]{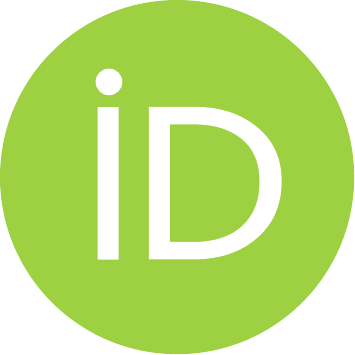}}
\newcommand{\orcidaffil}[1]{%
  \href{https://orcid.org/#1}{\usebox{\myorcidaffilbox}}}

\author{R. James Cotton$^{1,2,*}$\orcidaffil{0000-0001-5714-1400} , Anthony Cimorelli$^1$, Kunal Shah$^1$, Shawana Anarwala$^1$, \\ Scott Uhlrich$^3$, Tasos Karakostas$^{1,2}$ 
    \thanks{This work was generously supported by the Research Accelerator Program of the Shirley Ryan AbilityLab and with funding from the Restore Center P2C (NIH P2CHD101913).}
    \thanks{*rcotton@sralab.org, 1. Shirley Ryan AbilityLab, 2. Department of Physical Medicine and Rehabilitation,  Northwestern University, 3. Stanford University}
}

\begin{document}

\maketitle

\begin{abstract}
Markerless pose estimation allows reconstructing human movement from multiple synchronized and calibrated views, and has the potential to make movement analysis easy and quick, including gait analysis. This could enable much more frequent and quantitative characterization of gait impairments, allowing better monitoring of outcomes and responses to interventions. However, the impact of different keypoint detectors and reconstruction algorithms on markerless pose estimation accuracy has not been thoroughly evaluated. We tested these algorithmic choices on data acquired from a multicamera system from a heterogeneous sample of 25 individuals seen in a rehabilitation hospital. We found that using a top-down keypoint detector and reconstructing trajectories with an implicit function enabled accurate, smooth and anatomically plausible trajectories, with a noise in the step width estimates compared to a GaitRite walkway of only 8mm.
\end{abstract}

\section*{Introduction}\label{introduction}

Markerless pose estimation is an emerging approach for performing movement analysis, including gait analysis \cite{nakano_evaluation_2020, tang_comparison_2022, uhlrich_opencap_2022, kanko_concurrent_2020, pagnon_pose2sim_2021, kanko_assessment_2021, needham_development_2022, matthis_jonathan_samir_2022_7233714}. This technology is driven by advances in human pose estimation (HPE) \cite{zheng_deep_2020}. By acquiring images from multiple views with a calibrated camera system, and using HPE to localize joints in an image, the underlying 3D joint locations can be computed. While various algorithms can be employed for markerless pose estimation, the impact of these algorithmic choices has not been thoroughly evaluated. This study aims to address this gap in knowledge by evaluating the influence of the keypoint detection algorithm and joint location reconstruction algorithms on markerless pose estimation accuracy.

As a team involved in rehabilitation care, our overarching motivation is a system that allows easy, reliable, and routine movement analysis for clinical populations. This system would provide numerous benefits including better quantitative characterization of gait impairments and their response to interventions, free from the limitations of traditional optical capture systems which are restricted to laboratory use. For our goals, such a system must also produce clinically interpretable results, such as describing movements according to the International Society of Biomechanics standards \cite{wu_isb_2002}. However, obtaining good biomechanical fits with inverse kinematics to 3D joint locations requires high-quality trajectories. Optimizing our approach to obtain these high-quality trajectories from videos acquired with our multicamera acquisition system was the underlying motivation of this study.

We compared two different keypoint detectors. The first is OpenPose \cite{cao_openpose_2019}, which is a very popular keypoint detector that processes frames in a bottom-up fashion and locates the joints of all visible people in the scene. The second was a top-down algorithm from the MMPose library \cite{mmpose_contributors_openmmlab_2020}. Top-down algorithms only process the person of interest identified in a bounding box and generally have greater accuracy at the expense of having to process the bounding boxes \cite{zheng_deep_2020}.

We also compared three methods of reconstructing joint locations from detected keypoints. The first was a robust triangulation approach which uses the 2D observations of joints from each view to triangulate its 3D location. The second uses an optimization algorithm to find the best 3D joint locations that when reprojected through the camera model best align with the detected keypoints. This optimization approach allows additional constraints for smooth movement and consistent bone lengths. The third approach performs a similar optimization, but instead of directly optimizing the 3D joint locations, we optimize the parameters of an implicit function that maps from time to 3D pose. Implicit functions are widely used, for example in neural radiance fields for modeling 3D scenes \cite{gao_nerf_2022}, but to the best of our knowledge have not been applied to modeling movement trajectories.

To evaluate the accuracy of 3D reconstructions using these approaches, we compared our reconstructions of the heel and toe keypoints against measurements from a GaitRite walkway. This was tested on a heterogeneous convenience sample of ambulatory patients seen in a physical rehabilitation hospital who had a diverse range of gait patterns. We found that using a top-down keypoint detector and an implicit function to reconstruct the trajectories provided the best performance, with a standard deviation of the residual step width measurements of 8mm.

\section*{Methods}

\subsection*{Participants}

This study was approved by the Northwestern University Institutional Review Board. We recruited 25 participants from both the inpatient services and outpatient clinics at Shirley Ryan AbilityLab. This included people with a history of stroke (n=8), spinal tumor (n=1), traumatic brain injury (n=2), mild foot drop (n=1), knee osteoarthritis (n=2), and prosthetic users (n=11). Ages ranged from 27 to 78. Some participants used assistive devices including orthotics, rolling walkers, or a cane, and a few participants required contact guard assistance from someone nearby for safety. We intentionally recruited a diverse rehabilitation population both to ensure the system was robustly validated for all types of patients, and because this is the population we intend to apply this final system to. The data included 162 trials containing 2164 steps, with steps per participant ranging from 44-154, with over 1 million video frames.


\subsection*{Data Acquisition}

Multicamera data was collected with a custom system in a $7.4m \times 8m$ room with subjects walking the length of the diagonal ($11m$). We used 10 FLIR BlackFly S GigE cameras (and 8 in several early experiments), which were synchronized using the IEEE1558 protocol and acquired data at 30 fps, with a typical spread between timestamps of less than 100µs. We used a mixture of lenses including F1.4/6mm, F1.8/12m, F1.6/4.4-11mm with lens and positions selected to ensure at least three cameras covered the participants along the walkway, although the room geometry limited coverage in the corners.

The acquisition software was implemented in Python using the PySpin interface. For each experiment, calibration videos were acquired with a checkerboard ($7 \times 5$ grid of 110mm squares) spanning the acquisition volume. Extrinsic and intrinsic calibration was performed using the anipose library \cite{karashchuk_anipose_2020}. The intrinsic calibration included only the first distortion parameter. Foot contact and toe-off timing and location were acquired using a GaitRite walkway spanning the room diagonal \cite{mcdonough_validity_2001, bilney_concurrent_2003}. 

\subsubsection*{Video processing}

Our analysis pipeline was built upon our prior work with PosePipe  \cite{cotton_posepipe_2022}, which uses DataJoint \cite{yatsenko_datajoint_2015} to manage videos and the computational dependencies when running HPE. PosePipe supports both OpenPose \cite{cao_openpose_2019} and top-down algorithms from MMPose \cite{mmpose_contributors_openmmlab_2020}. Top-down algorithms require a bounding box to localize the person in order to compute the keypoints. PosePipe also allows using the bounding box to select the set of OpenPose keypoints corresponding to the person of interest.

\begin{figure}
\centering
\includegraphics[width=0.8\linewidth]{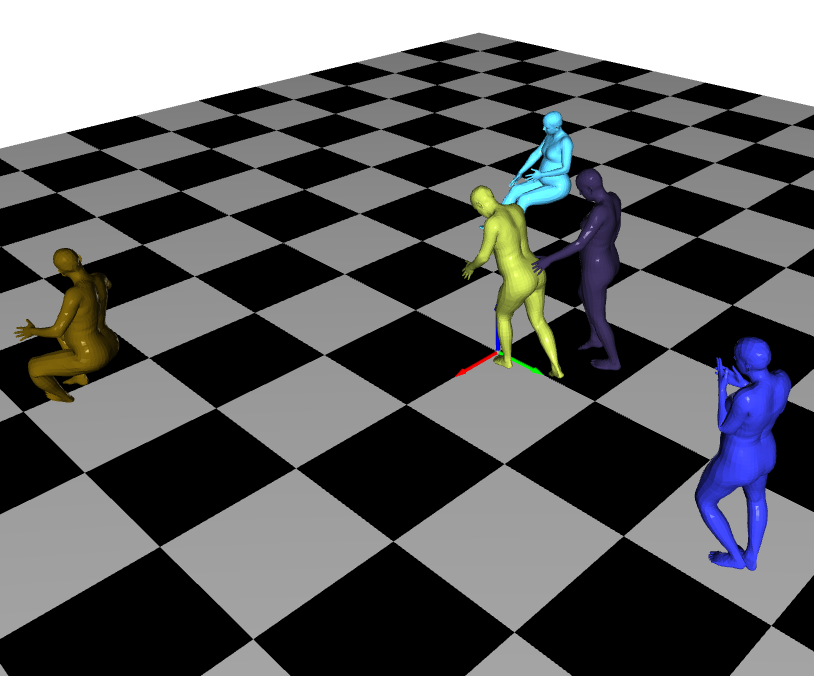}
\caption{Example visualization from a session using EasyMocap visualizer to identify the subject of interest (yellow), who is using a rolling walker and being assisted by a physical therapist. \label{fig:easymocap}}
\end{figure}

We developed an annotation tool using EasyMocap to identify the participant \cite{easymocap, dong2021fast}. EasyMocap takes the OpenPose outputs from all people seen in each camera and associates them with individuals across views and over time. 
We used the 3D visualization tool to identify the subject of interest from this reconstruction and then compute the bounding from those 3D joint locations by reprojecting them back into each camera view (Fig~\ref{fig:easymocap}).

Finally, we used these bounding boxes to obtain the keypoints using PosePipe. When using OpenPose, this simply involves selecting the keypoints for the appropriate person from each view. We ran OpenPose with both the default settings and a high resolution mode with network resolution for 1008 and 4 scales with a scale gap of 0.25. For MMPose, each image is cropped at the bounding box and this region is passed to a 2D keypoint detector. In this work, the specific architecture we use is an HRNet \cite{sun_deep_2019}  with a channel width of 48 that is pretrained on the Halpe dataset \cite{fang_alphapose_2022}. We selected this because, in contrast to the commonly used COCO dataset \cite{lin_microsoft_2014}, the Halpe dataset includes 136 keypoints, including heel and toe keypoints. 

We also attempted this analysis using the same MMPose HRNet algorithm trained on the COCO Wholebody dataset \cite{jin_whole_2020} but saw substantially worse performance in our initial investigations. This seemed to arise due to much more variable keypoints confidence estimates, and we did not pursue this analysis further.

\subsubsection*{Reconstruction}

We used three different approaches to reconstruct 3D joint locations from the detected 2D keypoints: robust triangulation, optimization of the 3D joint locations against a loss function that includes constraints for smoothness and skeleton consistency, and optimization against the same loss function using an implicit representation.

\paragraph{Robust triangulation}

Joint locations can be computed from 2D locations seen on multiple cameras that are spatially calibrated. The most common approach to this is the Direct Linear Transform (DLT) \cite{Hartley_Zisserman_2003}, which solves a series of linear equations from the camera projection matrices to find the point that minimizes the reprojection error. This can be further extended to weigh cameras by a keypoint confidence, $w_c$. For a point with observations in each camera $(u_c, v_c)$ where each camera has a projection matrix $P_c \in \mathbb R^{3 \times 4}$ with $\vec {p_c^i}$ indicating the $i^{th}$ row, we can create a series of linear equations:
\begin{equation}
\mathbf A = \left[ 
\begin{matrix}
w_1 \cdot (u_1 \vec p_1^{\, 3} - \vec p_1^{\, 1}) \\
w_1 \cdot (v_1 \vec p_1^{\, 3} - \vec p_1^{\, 2}) \\
\vdots \\
w_n \cdot (u_n \vec p_n^{\, 3} - \vec p_n^{\, 1}) \\
w_n \cdot (v_n \vec p_n^{\, 3} - \vec p_n^{\, 2}) \\
\end{matrix}
\right] \in \mathbb R^{2C \times 4}
\end{equation}
Where the optimal 3D location, $\mathbf x$, is the value that minimizes $||\mathbf A \mathbf x||$. This can be found by an SVD decomposition of A, $U \Sigma V=\mathbf A$. The last row of $V$ is proportional to $\hat x$ using homogeneous coordinates,  so  $\hat {\mathbf  x} = V[-1, :3] / V[-1, 3]$ (using Python notation). 

However, the DLT is sensitive to outliers, which will occur if another person occludes the view and has their joints detected, or if a joint is misdetected in one view, such as for people with limb differences like prosthetic users \cite{cimorelli_portable_2022}. A common approach to outlier rejection in 3D reconstruction is RANSAC, but this is a slow algorithm. Instead, we used a robust triangulation approach\cite{roy_triangulation_2022}. This robust approach determines the weight applied to each keypoint from each camera based on the geometric consistency of the triangulations from other cameras. We first perform triangulation from each pair of cameras to produce clusters of points in 3D, $\mathbf x^{c, c'}_{j,t}$, and in the equations below we omit explicit $j,t$ subscripts. Then, the geometric median location of each cluster is computed, 
\begin{equation*}
\tilde {\mathbf x}=\mathrm{GeoMed}(\{ \mathbf x^{c, c'} \quad \forall \quad (c, c') \in \left( \begin{matrix} N_c \\ 2 \end{matrix} \right) \} )
\end{equation*}, with $c$ and $c'$ indicating a pair of cameras from $C$. The distance from each point in the cluster to this median location is then computed $d^{c, c'} = \|\mathbf x^{c, c'} - \tilde {\mathbf x}\|$. A weight for each camera is then computed based on the distance of all the points triangulated using it,
\begin{equation*}
w_c=\mathrm{median}(\{ \exp(-\frac{d^{c, c'}}{\sigma^2}) \quad \forall \quad c'  \in C \backslash  c \})
\end{equation*}. We used a default $\sigma$ of 150mm. Any views with a confidence below the default threshold of $\gamma=0.5$ are excluded, so do not influence the cluster median. We also tested the influence of changing $\sigma$ and $\gamma$ (Table~\ref{table:full_hp}).

This robust triangulation is repeated for each joint and time point. To account for camera distortions, the 2D keypoints are first undistorted according to the intrinsic calibration parameters \cite{opencv_library}. The robust triangulation involves performing many SVD computations for each of the camera pairs (e.g., 45 pairs for 10 cameras) and the Halpe dataset includes 136 keypoints, thus requiring nearly 200,000 SVD operations per second of data. We implemented a custom camera library in Jax \cite{jax2018github}, including projections and triangulation, which allows parallelizing this on the GPU and kept the time to perform this triangulation on the order of seconds.

\paragraph{Optimization}
While robust triangulation is generally accurate and quick, it can produce implausible results such as inconsistent limb lengths and high-frequency noise. Post-processing can reduce these artifacts, but does not fully leverage the information available in the raw keypoints. Instead, we solve for the 3D joint locations that minimize a loss function including both the reprojection error and additional constraints including a smoothness loss and skeleton consistency loss.

For a 3D keypoint trajectory represented by $\mathbf X \in \mathbb R ^{T \times J \times 3}$ with $\mathbf x_{t,j}\in \mathbb R^3$ indicating the 3D location at a specific time and joint, we define three losses. The first is a reprojection loss:
$$
\mathcal L_{\Pi} = \frac{1}{T \cdot J \cdot C}\sum_{T, J, c\in C} w_{c,t,j} \, g \left( || \Pi_c \mathbf x_{t,j} - y_{t,j,c} || \right)
$$
where $\Pi_c$ is the projection operator for camera $c$ which also includes the non-linear intrinsic distortion. $y_{t,j,c} \in \mathbb R^2$ indicates the detected keypoint location for a given time point, joint and camera.  We use a Huber loss for $g(\cdot)$, which is quadratic within 5 pixels and then linear, as the Huber loss is more robust to outliers than MSE. The weights applied to Huber loss, $w$, are computed by the robust triangulation algorithm. We also experimented with using the weight terms from the keypoint confidences, and by making the Huber slope shallower after 10 pixels of error to be more robust to outliers (see Supplementary Materials).

The next term in the loss was a temporal smoothness loss, which penalizes the root mean squared change in the Euclidean 3D position over time:
$$\mathcal L_{\mathrm {smooth}} = \sqrt{\frac{1}{T \cdot J} \sum_{t=0 ... T-1 ,J} ||x_{t,j} - x_{t+1,j}||^2 }$$

The last term was a skeletal consistency loss. Let $l_{t,\vec{jj'}}$ be the length of a limb segment at time t in the skeleton, $\mathcal S$: 
\begin{equation*}
l_{t, \vec{jj'}} = ||x_{t,j}-x_{t,j'}|| \quad \forall \quad (j,j') \in \mathcal S
\end{equation*}
. We then create a skeleton consistency loss that is the root mean squared error of the limb length:
$$\mathcal L_{\mathrm {skeleton}} = \sqrt{\frac{1}{T \cdot J} \sum_{t=0,\, j}^T \left( l_{t, \vec{jj'}} - \bar{l_{\vec{jj'}}} \right) ^2}$$
For the set of pairs in the skeleton, we used the bilateral heel-toe, ankle-knee, knee-hip, shoulder-elbow, and elbow-wrist segments. We then define the final loss as:
$$\mathcal L = \mathcal L_\Pi + \lambda_1 \mathcal L_{\mathrm {smooth}} + \lambda_2 \mathcal L_{\mathrm {skeleton}}$$
with $\lambda_1=\lambda_2=0.1$. 

\paragraph{Trajectory Representation}
We compared two representations of the 3D keypoint trajectory when performing the optimization. The first is straightforward, simply storing the tensor $\mathbf X \in \mathbb R ^{T \times J \times 3}$ and directly minimizing this with respect to $\mathcal L$. 

The second representation used an implicit representation, $f_\theta : t \rightarrow \mathbf x_t \in \mathbb R^{J \times 3}$, where we learn the parameters of a multi-layer perceptron (MLP), $f_\theta$ to minimize $\mathcal L$. We used a five-layer MLP with increasing numbers of hidden units (128, 256, 512, 1024, 2048), each followed by a layer normalization and a $\mathrm {relu}$ non-linearity, followed by a final dense layer that mapped to the output size. We also used sinusoidal positional encoding \cite{vaswani_attention_2017} for $t$, which was scaled from 0 to $\pi$ over each trajectory. The positional encoding of time was concatenated to the output from each layer for the first four layers. We did not perform rigorous hyperparameter tuning of this architecture, but converged on this through some initial testing and visual inspection on a few trials.

The loss function and trajectories representations were implemented in Jax and both representations were optimized with Optax for 50,000 steps (for each trial) using an exponentially decaying learning rate with warmup from an initial learning rate of $1\times10^{-6}$ to a peak learning rate of $1\times10^{-4}$ and then decaying back to $1\times10^{-6}$.  It is notable that for a trial length of 30 seconds, the implicit representation has an order of magnitude more parameters in the MLP than the explicit representation, but given the efficiency of GPUs both approaches take less than a minute to fit a trajectory.

\paragraph{Geometric Consistency}
We also measured the geometric consistency between the reconstructed trajectories and the detected keypoints, based on the distance between the reprojected keypoints and the detected keypoints. To quantify this, we computed the fraction of the points below a threshold number of pixels, conditioned on being greater than a specified confidence interval. 
$$\delta_{t,j,c} = || \Pi_c \mathbf x_{t,j} - y_{t,j,c} ||$$
$$q(d,\lambda) = \frac{\sum (\delta_{t,j,c} < d)(w_{t,j,c} > \lambda)}{\sum w_{t,j,c} > \lambda}$$
For the primary metric of merit for reconstruction quality, we used $d=5$ pixels and $\lambda=0.5$. To make this comparable between OpenPose keypoints and MMPose, which has many more keypoints, we only computed this for the 25 MMPose keypoints that closely correspond to OpenPose keypoints.

\begin{figure*} 
\begin{center}
\includegraphics[width=0.95\linewidth]
{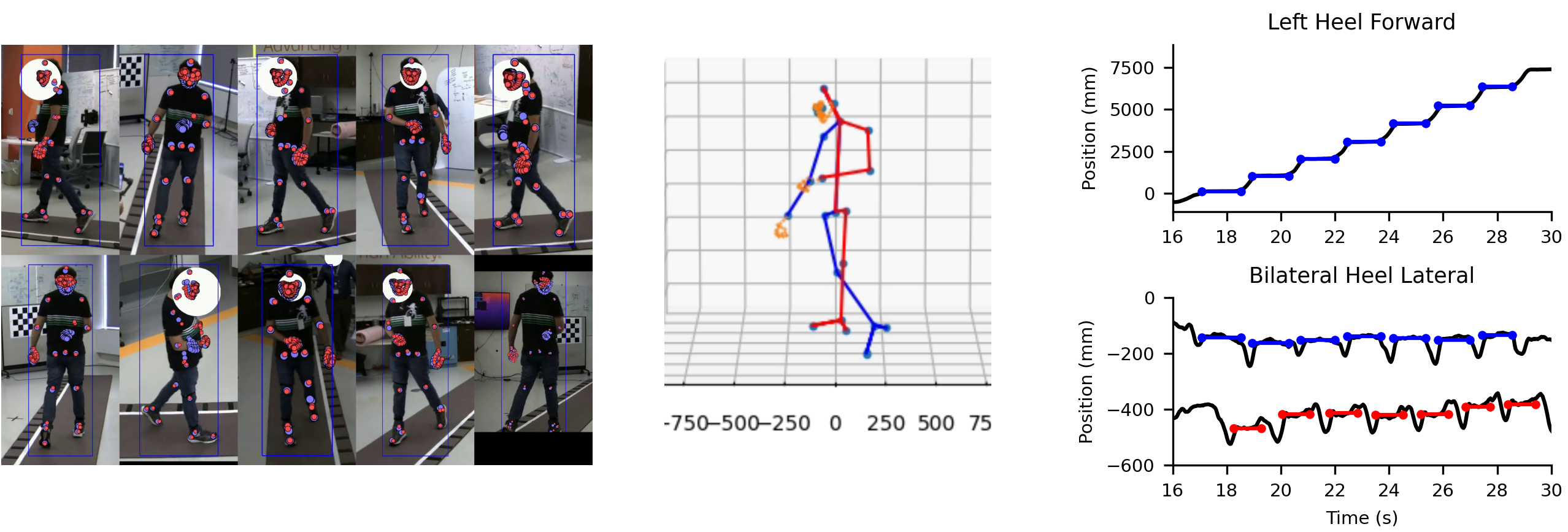}
\end{center}
\vspace{-1.5em}
\caption{Example reconstruction during walking. The left panel shows all of the views. The reconstructed joint locations are reprojected into the images as blue points with the detected keypoints as red points, showing good geometric alignment (we recommend zooming in on this figure). The middle panel shows the 3D reconstruction, showing even the hands and figures are well reconstructed. The right panel shows the left heel forward and lateral position with the position detected from the GaitRite marked in blue for the left foot and red for the right foot. Only the left foot is shown in the forward direction for visual clarity. \label{fig:visualizations}}
\vspace{-1.5em}
\end{figure*}

\paragraph{GaitRite comparison}
To test the spatial accuracy of our kinematic reconstructions, we compared these reconstructions against locations recorded from the GaitRite walkway. In order to directly compare these results, we had to calibrate the physical orientation of the GaitRite coordinate frame relative to the camera reference frame, which is fixed for a given recording session. This calibration also requires computing the temporal offset between the GaitRite data and the reconstructed kinematics. We performed this by first testing a range of possible temporal offsets for each session, computing the foot velocity when the GaitRite reported the foot was on the ground and then finding local minima that represent possible temporal offsets. We then iteratively searched for the combination of temporal offsets for each session and a rotation and translation for all sessions between the camera reference frame and GaitRite coordinates that minimized the difference between the estimates of heel and toe positions and those detected by the GaitRite. We also calibrated for a slight ($0.4-1.4\%$) difference in scale between the two systems.

After aligning our markerless coordinate frame to the GaitRite, we computed several error metrics, including the euclidean difference between each heel and toe position during the stance phase, and the step length and step width. The step width and length used the reconstructed heel kinematics. We compared it to the ``Base of Support'' field exported by GaitRite, which corresponds to the distance between the heel contact position and the line of progression from the contralateral foot. We computed the residuals of these errors per step over all the trials and report aggregate statistics on these errors.

\section*{Results}

\subsection*{System usability and annotation}

We found the system generally easy to use and encountered few technical challenges. 
Setup time was minimal given the markerless nature of the acquisition. The annotation tool we developed using EasyMocap allowed us to separate the participant from others in the room, even when up to six people were present, although there was occasional noise in the EasyMocap reconstructions at the edge of the acquisition volume. 
Even when an additional team member was walking beside the participant for stabilization, separation worked well in the volume. The processing pipeline using PosePipe and built upon DataJoint also made it easy to manage analysis and results from the 1000s of acquired videos.

\subsection*{Qualitative Results}

We created videos showing both the 3D reconstructions, the keypoints reprojected onto the videos from each view, and traces comparing the trajectories with the GaitRite data overlaid as in Figure~\ref{fig:visualizations}. We manually reviewed many trials for different algorithm variations. We found that the MMPose keypoints were both better aligned to the real joints and resulted in better-aligned reprojected joint locations. The optimized representations also had much less jitter and anatomically implausible transitions compared to robust triangulation. This was apparent on the foot traces, with robust triangulation having more noise when there were fewer views tracking the points (i.e., at the edge of the volume). At times, this prevented us from computing errors when the foot location was not well defined, which was particularly common when using OpenPose and robust triangulation.

\subsection*{Geometric Consistency}

\begin{figure}
\centering
\includegraphics[width=0.5\linewidth]{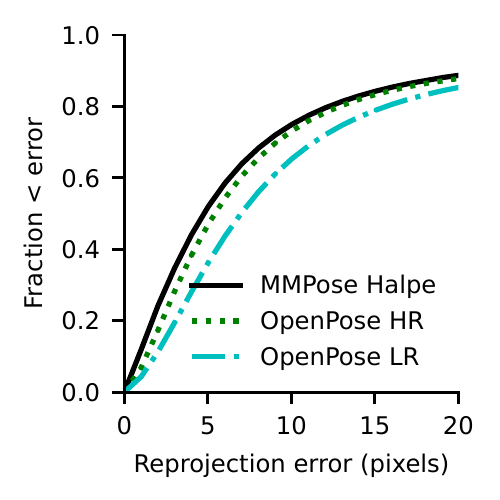}
\vspace{-1.5em}
\caption{Geometric consistency between the reprojection of 3D points and the 2D detection locations. This compares reconstructions using OpenPose and MMPose, with the x-axis indicating the threshold number of pixels and the y-axis showing the fraction of points within this threshold distance. The curve only includes 2D keypoints with a confidence greater than 0.5. Reconstructions use the robust triangulation.\label{fig:geometric_consistency}}
\vspace{-1.5em}
\end{figure}

We first compared the reprojection quality metric between reconstructions using keypoints from OpenPose HR and LR versus MMPose with the Halpe dataset, using robust triangulation. Fig~\ref{fig:geometric_consistency}a shows the fraction of reprojected pixels within a given threshold when selecting keypoints with a confidence greater than 0.5 and demonstrates that MMPose produces significantly greater geometric consistency, although OpenPose HR is close. We report this in Table~\ref{table:gaitrite}, where $GC_x=q(x, 0.5)$ indicates the fraction of pixels that reproject within $x$ pixels of error, for keypoints with a confidence greater that 0.5. These differences were statistically significant at 5 pxels ($p < 1.0\times10^{-5}$ for Kruskal Wallis test and post-hoc per-group test differences). 

In contrast, we did not see much difference in the geometric consistency for the MMPose keypoints when changing the reconstruction algorithm between robust triangulation and the two optimization-based approaches. It also did not change markedly when altering the $\sigma$ and $\gamma$ parameters for the robust triangulation, or when using the keypoint confidences as weights in the reprojection loss (Table~\ref{table:full_hp}). 

\subsection*{Skeletal consistency and smoothness}

The motivation for the optimization approach is to ensure the inferred 3D location trajectories are physically plausible, including not having high-frequency jitter and producing consistent bone lengths during a trajectory. We compared the  $\mathcal L_{\mathrm {smooth}}$ and $\mathcal L_{\mathrm {skeleton}}$ for each of the three methods (Table~\ref{table:gaitrite}). For both metrics, the implicit representation had the best performance and robust triangulation had the worst performance ($p < 1.0\times10^{-10}$ for Kruskal Wallis test and for post-hoc per-group test differences on MMPose outputs).

\subsection*{GaitRite comparison}

We compared the location of the heel and toe position from the different keypoints and reconstruction approaches against the positions reported from the GaitRite walkway. Table~\ref{table:gaitrite} reports the normalized IQR, $\sigma_{IQR}=0.7413 \cdot IQR(x)$, of the differences between the GaitRite parameters and those from our kinematic tracking of the feet. We use this metric as it is robust to a few outliers steps that occur for some methods. Mean and std are also reported in Table~\ref{table:full_rmse}.

The most notable trend was that using keypoints from MMPose resulted in substantially lower errors than both OpenPose variants. With OpenPose and robust triangulation, it was also not uncommon to have large errors near the end of the walkway, particularly if a keypoint was jittering near the confidence threshold and tracking was intermittently lost. When there was too much noise in the trace or tracking was lost, steps were discarded, which was more favorable for OpenPose. When comparing the specific reconstruction methods applied to MMPose keypoints, we saw less of an influence of the specific reconstruction algorithm, with only a few mm difference for most metrics between the approaches. Implicit optimization trended towards a small but slight advantage. 

\begin{figure}
\centering
\includegraphics[width=1.0\linewidth]{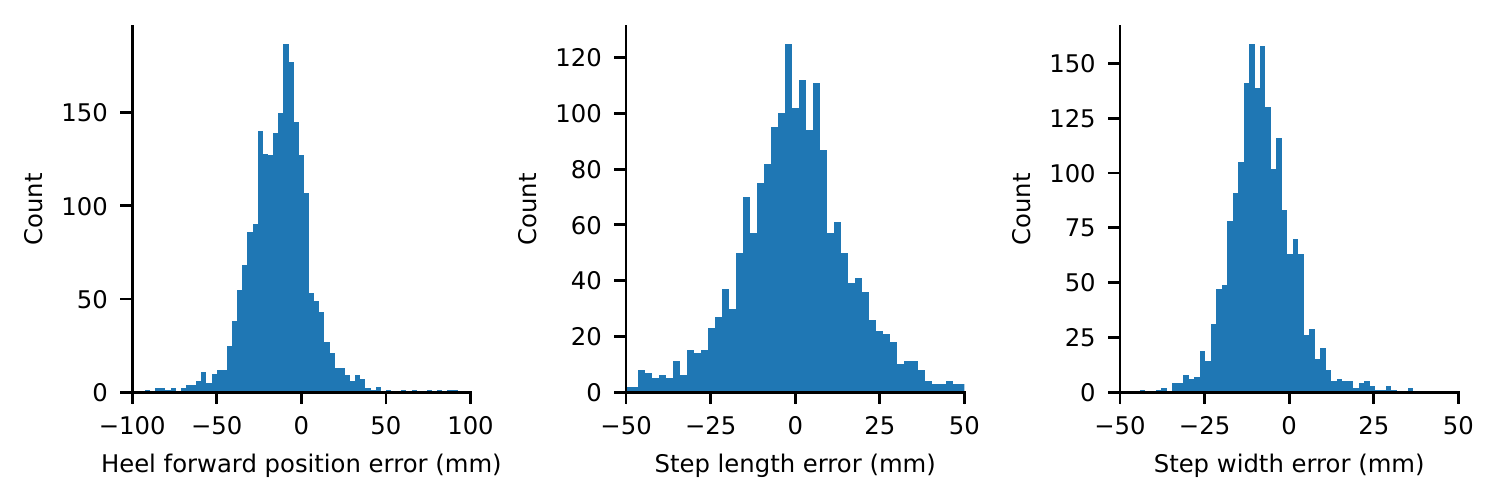}
\vspace{-1.5em}
\caption{Histogram of errors when reconstructing gait with the optimized implicit representation. The left panel shows the forward step position errors, which shows the bias between HPE and GaitRite position. The middle panel shows the histogram of step length errors, which is centered because the bias is constant between steps. Step width also shows a bias.\label{fig:gaitrite_errors}}
\vspace{-1.5em}
\end{figure}

The GaitRite walkway detects pressure whereas markerless pose estimation uses visual cues, so it is perhaps not surprising if there is a difference between the location of the two. When computing the residuals in the forward direction, we reversed the sign of the error when the participant was walking in the descending direction on the walkway, which keeps the relative offset invariant to the walking direction. Fig~\ref{fig:gaitrite_errors} demonstrates this difference between modalities with a mean offset of 13mm in the histogram for the foot forward position error. When measuring step and stride length, the bias cancels out (Fig~\ref{fig:gaitrite_errors}). However, this is not the case for step width, since a lateral offset between modalities will be doubled when taking the difference between foot positions and we see a bias of -8mm. The $\sigma_{IQR}$ metric in Table~\ref{table:gaitrite} is insensitive to these biases and captures the clinically relevant aspect where we typically want to measure changes in gait parameters and a few mm offset in the absolute value is not clinically meaningful.

\setlength{\tabcolsep}{2pt}
\begin{table*}[t]
  \caption{\label{table:gaitrite}}
  \vspace{-1.5em}
  \centering
  \begin{tabular}{llrrrrrrrrrrrrr}
\toprule
 &  & $GC_5$ & $GC_{10}$ & $GC_{20}$ & $\mathcal L_\Pi$ & $\mathcal L_{\mathrm{skeleton}}$ & $\mathcal L_{\mathrm{smooth}}$ & Heel X & Heel Y & Toe X & Toe Y & Step Len & Stride Len & Step Width \\
Keypoints & Method &  &  &  &  &  &  & (mm) & (mm) & (mm) & (mm) & (mm) & (mm) & (mm) \\
\midrule
\multirow[c]{3}{*}{MMPoseHalpe} & Implicit Optimization & \bfseries 0.54 & \bfseries 0.77 & \bfseries 0.90 & 9.5 & \bfseries 0.3 & 18.9 & \bfseries 15.8 & \bfseries 9.6 & \bfseries 12.6 & \bfseries 15.3 & \bfseries 13.2 & \bfseries 9.6 & \bfseries 8.0 \\
 & Explicit Optimization & 0.53 & 0.75 & 0.88 & 10.3 & 27.6 & 37.2 & 20.9 & 10.6 & 15.4 & 16.7 & 14.8 & 10.0 & 8.6 \\
 & Robust Triangulation & 0.53 & 0.76 & 0.90 & 11.9 & 138.1 & 71.4 & 17.6 & 9.7 & 14.7 & 16.4 & 14.5 & 10.8 & 8.5 \\
\midrule
\multirow[c]{3}{*}{OpenPose LR} & Implicit Optimization & 0.37 & 0.64 & 0.86 & 13.2 & 14.2 & 19.2 & 28.2 & 13.9 & 42.0 & 24.0 & 27.5 & 19.6 & 14.1 \\
 & Explicit Optimization & 0.35 & 0.63 & 0.84 & 15.2 & 73.6 & 70.4 & 27.0 & 14.3 & 33.4 & 24.8 & 18.8 & 15.8 & 13.6 \\
 & Robust Triangulation & 0.36 & 0.66 & 0.87 & 15.0 & 1097.7 & 123.1 & 19.6 & 15.0 & 24.8 & 26.4 & 18.0 & 13.7 & 12.2 \\
\midrule
\multirow[c]{3}{*}{OpenPose HR} & Implicit Optimization & 0.48 & 0.74 & 0.89 & \bfseries 9.3 & 14.2 & \bfseries 17.6 & 26.3 & 13.2 & 41.5 & 22.6 & 29.1 & 17.3 & 13.0 \\
 & Explicit Optimization & 0.46 & 0.73 & 0.89 & 9.9 & 32.7 & 41.9 & 23.7 & 13.2 & 36.5 & 22.5 & 18.5 & 14.5 & 12.5 \\
 & Robust Triangulation & 0.47 & 0.74 & 0.89 & 12.7 & 1251.6 & 124.0 & 19.8 & 14.5 & 23.3 & 23.2 & 17.2 & 12.4 & 11.0 \\
\bottomrule
\end{tabular}
  \vspace{-1.5em}
\end{table*}

\subsection*{Hyperparameters and variations}

There are a number of hyperparameters including what keypoint confidence to threshold at, $\gamma$, and for robust triangulation what distance to use for $\sigma$. For example we found that OpenPose tended to produce lower confidence values so we explored setting $\gamma=0.3$ instead of the default of 0.5, which did improve some metrics but not to the level of using MMPose. Robust triangulation generally performed worse with smaller values of $\sigma$. See Appendix Table~\ref{table:full_hp}.

\section*{Discussion}\label{discussion}

Our strongest finding was that using 2D keypoints from MMPose produced substantially better results than using OpenPose measured both by the consistency with reprojecting 3D positions back into the camera frame and when compared to the GaitRite. This is not surprising given that top-down methods typically outperform bottom-up methods for 2D detection accuracy \cite{zheng_deep_2020}, but is worth highlighting given that OpenPose is still used frequently, even in recent works analyzing walking. However, top-down approaches introduce additional complexities including computing and using bounding boxes for the person of interest. This challenge was mitigated through our previous development of PosePipe \cite{cotton_posepipe_2022} to facilitate using cutting edge HPE algorithms and further by developing an annotation system based on EasyMocap \cite{easymocap}. While we did not focus on it for this study, we found that after triangulating from the MMPose Halpe keypoints, we could reconstruct individual finger joints and will quantify this hand tracking performance in future studies.

The reconstruction method has a significant influence on plausibility. The optimization based approaches, by the design of the loss function, produce much more anatomically and kinematically realistic results and are less inclined to have large noise events when keypoints are noisy. We were motivated to develop the implicit representation for optimizing trajectories after our experience in prior work \cite{cotton_kinematic_2020} that the addition of constraints such as smoothness or constant body shape seems to make optimization fairly slow and difficult to fully converge with gradient descent. We speculate this is because it takes many optimization iterations for changes to equilibrate through the entire sequence. We predicted that because the effects of parameters in the implicit function are non-local, it would be better able to account for these constraints. Our results showing that the implicit representation had better performance for each of the loss terms despite a large number of optimization steps support this. The high-performance implementation of this optimization algorithm in Jax also means these reconstructions can be performed rather quickly, which we previously found prohibitive.

A planned next step for this system is to perform biomechanical fits to the resulting motion (e.g., \cite{uhlrich_opencap_2022}). Biomechanical analysis with inverse kinematics can be sensitive to outliers in marker locations and so we anticipate the more plausible trajectories found with the implicit optimization will improve these results. An additional benefit of the implicit representation is that because it learns a \emph{function} that maps from time to pose, it inherently supports sampling at arbitrary timesteps. This is a useful benefit when optimizing trajectories against multimodal data like sensors and video \cite{cotton_kinematic_2020}.

This work has numerous limitations and opportunities for improvement.
We found greater performance benefits from improving the quality of the 2D keypoints than from the reconstruction algorithm.
The geometric consistency curves show that only half of the reprojected points are within 5 pixels but 90\% are within 20 pixels, suggesting there is further room for improvement.
This includes having the 2D keypoint locations more consistently project to a fixed internal location. For example, we noticed that the detected hip locations become biased upward when looking down at an individual. 
We note that the robust triangulation approach was described in a paper on multiview self-supervised learning (SSL)\cite{roy_triangulation_2022}, where the information between views is used to improve the geometric consistency between 2D keypoint detectors, and we plan to implement this for fine-tuning keypoint detectors.
This would have the added benefit of allowing learning from a diverse clinical population, which can address limitations we have previously noted such as tracking the location of prosthetic limbs \cite{cimorelli_portable_2022}.
Keypoint detectors would also be improved for biomechanics by learning denser keypoints over the trunk, as their absence can limit understanding pelvis movement and reconstructing hip angles without training additional models to mitigate this \cite{uhlrich_opencap_2022}.



Another limitations of our pipeline is that, while EasyMocap works quite well and allowed us to annotate these videos orders of magnitude faster than with our prior PosePipe tool, many of our errors were the result of poor bounding box localization when subjects were at the edge of the recording volume. This is likely attributable both to noise in the OpenPose keypoints used to perform the initial reconstruction and because our room was not large enough to have cameras filming from diverse perspectives to better constrain the geometry. While this wasn't a problem in the gait acquisition volume, we expect it could still be mitigated using recent multiview fusion approaches \cite{dong_shape-aware_2021,wang_direct_2021}. The $1\%$ scaling error seen between the GaitRite measurements and our system also indicate the need to improve the calibration routine.

Finally, for any clinical application it is important to have confidence measures, including for the tracking accuracy. For example, we recently described a lifting algorithm that produces well calibrated distributions of 3D joint locations \cite{pierzchlewicz_multi-hypothesis_2022}. While the trajectory optimization-based approaches are more robust to occlusions and keypoint noise, they perform this partly by extrapolating. The geometric consistency measure and the weights from the robust triangulation algorithm both provide measures of reconstruction quality, but further work will be required to map these to calibrated confidence estimates. In future work, we anticipate performing this while also investigating the influence of the number and geometry of views on reconstruction accuracy.

In conclusion, we found acquiring gait data with our synchronized multicamera system reliable and easy to perform. The minimal setup time and speed at which data can be collected made it feasible for us to easily recruit participants seen at our rehabilitation facility as both inpatients and outpatients and quickly obtain quantiative gait data. Reconstruction using MMPose Halpe keypoints with implicit trajectories produced the most accurate results with step width and length noise of under 10mm.




\printbibliography


\newpage
\newpage

\setcounter{table}{0}
\renewcommand{\thetable}{A\arabic{table}}

\begin{table*}
  \caption{Full table of results from hyperparameter exploration. \label{table:full_hp}}
  \centering
  \tiny
  \begin{tabular}{llrrrrrrrrrrrrr}
\toprule
 &  & $GC_5$ & $GC_{10}$ & $GC_{20}$ & $\mathcal L_\Pi$ & $\mathcal L_{\mathrm{skeleton}}$ & $\mathcal L_{\mathrm{smooth}}$ & Heel Forward & Heel Lateral & Toe Forward & Toe Lateral & Step Length & Stride Length & Step Width \\
Keypoints & Method &  &  &  &  &  &  & (mm) & (mm) & (mm) & (mm) & (mm) & (mm) & (mm) \\
\midrule
\multirow[c]{12}{*}{MMPoseHalpe} & Explicit Optimization & 0.53 & 0.75 & 0.87 & 11.22 & 38.71 & 44.42 & 20.9 & 10.6 & 15.4 & 16.7 & 14.8 & 10.0 & 8.6 \\
 & Explicit Optimization KP Conf, MaxHuber=10 & 0.56 & 0.72 & 0.87 & 12.80 & 4.31 & 29.64 & 15.6 & 9.9 & 13.3 & \bfseries 14.8 & 13.7 & 9.0 & 8.8 \\
 & Implicit Optimization & 0.54 & 0.76 & 0.89 & 9.87 & 0.43 & 18.93 & 15.8 & \bfseries 9.6 & \bfseries 12.6 & 15.3 & 13.2 & 9.6 & \bfseries 8.0 \\
 & Implicit Optimization $\gamma=0.3$ & 0.51 & 0.72 & 0.85 & 12.98 & 59.58 & 57.56 & 22.5 & 11.0 & 17.8 & 17.8 & 17.3 & 10.0 & 8.6 \\
 & Implicit Optimization KP Conf & 0.56 & \bfseries 0.77 & 0.89 & 9.04 & 0.47 & 19.30 & 16.3 & 9.8 & 13.0 & 15.0 & 15.0 & 10.9 & 8.1 \\
 & Implicit Optimization KP Conf, MaxHuber=10 & \bfseries 0.58 & 0.73 & 0.87 & 9.59 & 0.25 & 17.56 & 15.8 & 10.0 & 13.4 & 15.4 & 14.6 & \bfseries 9.0 & 8.8 \\
 & Implicit Optimization, MaxHuber=10 & 0.55 & 0.72 & 0.87 & 10.27 & 0.27 & 17.27 & 15.6 & 10.3 & 13.1 & 15.2 & 13.8 & 9.3 & 9.2 \\
 & Robust Triangulation & 0.52 & 0.75 & 0.89 & 948.60 & 1190.87 & 123.90 & 17.6 & 9.7 & 14.7 & 16.4 & 14.5 & 10.8 & 8.5 \\
 & Robust Triangulation $\gamma=0.3$ & 0.50 & 0.74 & 0.88 & 13599.49 & 7454.23 & 61728583.86 & 18.0 & 9.8 & 15.1 & 16.5 & 14.5 & 11.3 & 8.7 \\
 & Robust Triangulation $\sigma=100$ & 0.54 & 0.76 & 0.89 & 224269.25 & 5792.93 & 61728533.61 & 17.0 & 9.8 & 14.3 & 16.1 & 13.4 & 10.2 & 8.4 \\
 & Robust Triangulation $\sigma=50$ & 0.56 & 0.76 & 0.88 & 24065102.26 & 7025238.76 & 370435608.42 & 17.0 & 9.7 & 14.2 & 15.8 & \bfseries 12.9 & 9.5 & 8.5 \\
 & Triangulation & 0.49 & 0.69 & 0.85 & 53.53 & 38892.65 & 3765443945.39 & 24.3 & 13.1 & 19.5 & 17.8 & 22.0 & 21.3 & 10.5 \\
\midrule 
\multirow[c]{12}{*}{OpenPose HR} & Explicit Optimization & 0.46 & 0.72 & 0.88 & 10.23 & 45.31 & 44.79 & 23.7 & 13.2 & 36.5 & 22.5 & 18.5 & 14.5 & 12.5 \\
 & Explicit Optimization KP Conf, MaxHuber=10 & 0.40 & 0.58 & 0.76 & 15.15 & 54.52 & 72.59 & 67.3 & 20.0 & 69.7 & 24.9 & 309.5 & 33.4 & 161.0 \\
 & Implicit Optimization & 0.48 & 0.72 & 0.88 & 9.43 & 15.08 & 18.28 & 26.3 & 13.2 & 41.5 & 22.6 & 29.1 & 17.3 & 13.0 \\
 & Implicit Optimization $\gamma=0.3$ & 0.45 & 0.72 & 0.88 & 10.55 & 71.73 & 58.87 & 25.0 & 13.9 & 37.7 & 23.3 & 16.8 & 13.1 & 11.7 \\
 & Implicit Optimization KP Conf & 0.51 & 0.76 & 0.89 & \bfseries 8.55 & 10.06 & 18.47 & 32.3 & 14.2 & 53.3 & 25.4 & 46.5 & 19.1 & 12.8 \\
 & Implicit Optimization KP Conf, MaxHuber=10 & 0.49 & 0.68 & 0.85 & 10.01 & \bfseries 0.21 & 16.57 & 43.7 & 16.5 & 63.3 & 27.7 & 90.2 & 24.3 & 17.0 \\
 & Implicit Optimization, MaxHuber=10 & 0.44 & 0.65 & 0.83 & 11.03 & 0.22 & \bfseries 16.34 & 40.8 & 15.9 & 53.8 & 25.3 & 76.3 & 23.9 & 18.6 \\
 & Robust Triangulation & 0.47 & 0.73 & 0.88 & 16.13 & 2875.21 & 159.23 & 19.8 & 14.5 & 23.3 & 23.2 & 17.2 & 12.4 & 11.0 \\
 & Robust Triangulation $\gamma=0.3$ & 0.46 & 0.74 & \bfseries 0.90 & 12.23 & 4385.39 & 135.26 & 18.3 & 14.2 & 21.4 & 23.4 & 16.9 & 11.7 & 10.8 \\
 & Robust Triangulation $\sigma=100$ & 0.48 & 0.74 & 0.88 & 13.74 & 3187.83 & 164.64 & 19.4 & 14.3 & 22.2 & 23.4 & 16.7 & 12.4 & 11.1 \\
 & Robust Triangulation $\sigma=50$ & 0.50 & 0.74 & 0.87 & 328.18 & 13335.24 & 123457099.78 & 19.3 & 14.3 & 22.0 & 23.0 & 16.1 & 11.7 & 10.9 \\
 & Triangulation & 0.49 & 0.73 & 0.87 & 11.24 & 16746733.00 & 5247223290.94 & 25.1 & 15.8 & 27.8 & 23.4 & 20.5 & 15.5 & 12.5 \\
\midrule 
\multirow[c]{12}{*}{OpenPose LR} & Explicit Optimization & 0.36 & 0.63 & 0.82 & 16.13 & 86.40 & 75.11 & 27.0 & 14.3 & 33.4 & 24.8 & 18.8 & 15.8 & 13.6 \\
 & Explicit Optimization KP Conf, MaxHuber=10 & 0.33 & 0.53 & 0.75 & 17.96 & 30.18 & 61.58 & 50.4 & 18.1 & 63.2 & 27.6 & 135.8 & 28.6 & 48.5 \\
 & Implicit Optimization & 0.38 & 0.64 & 0.84 & 13.23 & 14.48 & 19.44 & 28.2 & 13.9 & 42.0 & 24.0 & 27.5 & 19.6 & 14.1 \\
 & Implicit Optimization $\gamma=0.3$ & 0.35 & 0.62 & 0.82 & 24.19 & 105.09 & 84.02 & 27.1 & 15.2 & 38.1 & 25.1 & 19.1 & 14.9 & 13.3 \\
 & Implicit Optimization KP Conf & 0.41 & 0.68 & 0.87 & 11.95 & 13.03 & 19.20 & 30.5 & 14.6 & 48.9 & 26.7 & 35.4 & 20.8 & 13.1 \\
 & Implicit Optimization KP Conf, MaxHuber=10 & 0.38 & 0.60 & 0.82 & 13.76 & 0.22 & 16.81 & 43.2 & 17.2 & 59.6 & 27.9 & 88.0 & 26.8 & 18.7 \\
 & Implicit Optimization, MaxHuber=10 & 0.35 & 0.56 & 0.80 & 14.99 & 0.22 & 16.59 & 38.9 & 16.5 & 53.1 & 25.0 & 73.1 & 27.0 & 19.5 \\
 & Robust Triangulation & 0.36 & 0.65 & 0.85 & 35.22 & 4464.95 & 193.42 & 19.6 & 15.0 & 24.8 & 26.4 & 18.0 & 13.7 & 12.2 \\
 & Robust Triangulation $\gamma=0.3$ & 0.36 & 0.65 & 0.86 & 36.43 & 4034.41 & 161.17 & 19.5 & 14.8 & 24.6 & 25.9 & 18.6 & 13.5 & 12.0 \\
 & Robust Triangulation $\sigma=100$ & 0.37 & 0.66 & 0.85 & 104.17 & 5347.91 & 208.02 & 19.6 & 14.8 & 24.0 & 26.6 & 17.7 & 13.5 & 12.1 \\
 & Robust Triangulation $\sigma=50$ & 0.39 & 0.66 & 0.85 & 6103.90 & 63692.72 & 852.62 & 20.2 & 14.5 & 24.2 & 25.7 & 17.4 & 13.6 & 12.3 \\
 & Triangulation & 0.35 & 0.64 & 0.83 & 15.95 & 2676075.92 & 3518557850.82 & 25.2 & 16.8 & 29.0 & 25.8 & 21.3 & 17.4 & 13.6 \\
\bottomrule
\end{tabular}
\end{table*}

\begin{table*}
  \tiny
  \caption{The mean and root mean squared error of the residuals for all the gait parameters for all of the methods tested in mm. This is in contrast to Table~\ref{table:gaitrite}, which reports the $\sigma_{IQR}$ of the residuals. Because of the constant difference between the definition of heel and toe location for computer vision and pressure-based measurements, the distribution of errors are biased (e.g. Fig~\ref{fig:gaitrite_errors}) and these performance numbers are slightly worse. \label{table:full_rmse}}
  \centering
  \textbf{Mean}\\
\vspace{1em}
\begin{tabular}{llrrrrrrr}
\toprule
 &  & Heel Forward & Heel Lateral & Toe Forward & Toe Lateral & Step Length & Stride Length & Step Width \\
Keypoints & Method &  &  &  &  &  &  &  \\
\midrule
\midrule
\multirow[c]{12}{*}{MMPoseHalpe} & Explicit Optimization & -19.9 & 3.8 & -2.4 & 3.0 & 0.5 & -2.8 & -8.3 \\
 & Explicit Optimization KP Conf, MaxHuber=10 & -13.1 & -0.6 & 4.0 & -1.7 & -1.0 & -3.5 & -9.0 \\
 & Implicit Optimization & -12.5 & -1.6 & 1.6 & -2.6 & 0.2 & -1.6 & -7.9 \\
 & Implicit Optimization $\gamma=0.3$ & -21.2 & 6.1 & -1.0 & 5.6 & -0.7 & -5.3 & -8.0 \\
 & Implicit Optimization KP Conf & -12.5 & -0.6 & 2.2 & -1.7 & -0.5 & -4.5 & -8.3 \\
 & Implicit Optimization KP Conf, MaxHuber=10 & -13.0 & -0.6 & 3.2 & -2.0 & -1.3 & -3.3 & -9.1 \\
 & Implicit Optimization, MaxHuber=10 & -15.3 & -1.8 & 4.7 & -2.9 & 0.4 & -1.6 & -8.7 \\
 & Robust Triangulation & -11.1 & -0.8 & -0.4 & -1.9 & 0.2 & -1.8 & -8.4 \\
 & Robust Triangulation $\gamma=0.3$ & -10.0 & -0.5 & -0.6 & -1.3 & 0.0 & -2.6 & -8.5 \\
 & Robust Triangulation $\sigma=100$ & -11.2 & -1.0 & 0.5 & -2.0 & 0.3 & -1.3 & -8.6 \\
 & Robust Triangulation $\sigma=50$ & -10.7 & -0.8 & 1.9 & -2.3 & 0.1 & -1.5 & -8.9 \\
 & Triangulation & -7.8 & 1.4 & 1.2 & -0.2 & -2.8 & -6.6 & -6.9 \\
 \midrule
\multirow[c]{12}{*}{OpenPose HR} & Explicit Optimization & 8.2 & -0.6 & -16.9 & -0.0 & -0.2 & -4.1 & -11.9 \\
 & Explicit Optimization KP Conf, MaxHuber=10 & 50.9 & 10.7 & -57.7 & 8.7 & 208.6 & 3.0 & 177.7 \\
 & Implicit Optimization & 24.0 & -4.5 & -18.5 & -3.9 & 0.3 & -1.5 & -13.3 \\
 & Implicit Optimization $\gamma=0.3$ & 3.0 & 4.4 & -19.5 & 5.5 & -0.5 & -3.4 & -12.0 \\
 & Implicit Optimization KP Conf & 32.9 & -2.3 & -25.1 & -0.7 & 1.2 & -1.0 & -11.8 \\
 & Implicit Optimization KP Conf, MaxHuber=10 & 43.5 & -2.4 & -29.7 & -0.3 & -0.4 & -4.5 & -13.9 \\
 & Implicit Optimization, MaxHuber=10 & 42.5 & -4.0 & -24.5 & -3.0 & -0.7 & -3.5 & -15.0 \\
 & Robust Triangulation & 2.1 & -2.6 & -0.3 & 0.8 & 1.0 & 5.6 & -14.0 \\
 & Robust Triangulation $\gamma=0.3$ & -0.7 & -2.4 & -1.5 & -2.4 & 0.8 & 0.8 & -15.2 \\
 & Robust Triangulation $\sigma=100$ & 1.9 & -2.5 & 0.7 & 0.9 & 0.9 & 5.9 & -14.4 \\
 & Robust Triangulation $\sigma=50$ & 2.0 & -2.5 & 2.3 & 0.6 & 1.2 & 6.3 & -14.4 \\
 & Triangulation & 0.9 & -1.3 & 3.7 & -1.8 & 3.1 & 6.1 & -11.2 \\
 \midrule
\multirow[c]{12}{*}{OpenPose LR} & Explicit Optimization & 6.7 & 5.2 & -14.7 & 4.6 & -0.8 & -7.1 & -11.1 \\
 & Explicit Optimization KP Conf, MaxHuber=10 & 47.8 & -1.9 & -50.9 & 5.7 & 91.7 & 0.2 & 70.2 \\
 & Implicit Optimization & 25.4 & -4.2 & -22.4 & -3.6 & 0.5 & -0.8 & -12.8 \\
 & Implicit Optimization $\gamma=0.3$ & 4.4 & 8.4 & -18.5 & 9.1 & -0.4 & -6.5 & -11.1 \\
 & Implicit Optimization KP Conf & 31.5 & -2.8 & -25.2 & -1.2 & 1.4 & 0.0 & -12.3 \\
 & Implicit Optimization KP Conf, MaxHuber=10 & 46.9 & -3.0 & -29.8 & -1.0 & -0.1 & -3.2 & -14.7 \\
 & Implicit Optimization, MaxHuber=10 & 42.5 & -4.5 & -27.7 & -3.1 & 0.3 & -2.0 & -13.8 \\
 & Robust Triangulation & -1.8 & -2.7 & -5.7 & 0.5 & 1.3 & 5.6 & -15.2 \\
 & Robust Triangulation $\gamma=0.3$ & -1.9 & -2.7 & -6.9 & -2.5 & 1.1 & 1.3 & -14.9 \\
 & Robust Triangulation $\sigma=100$ & -2.2 & -2.2 & -4.4 & 0.5 & 1.3 & 5.5 & -15.0 \\
 & Robust Triangulation $\sigma=50$ & -3.8 & -3.0 & -4.7 & 0.5 & 1.2 & 5.3 & -14.9 \\
 & Triangulation & 0.1 & -1.1 & -1.2 & -0.8 & 2.7 & 5.4 & -12.5 \\
\bottomrule
\end{tabular}

\vspace{2em}
\textbf{Std}\\
\vspace{1em}
\begin{tabular}{llrrrrrrr}
\toprule
 &  & Heel Forward & Heel Lateral & Toe Forward & Toe Lateral & Step Length & Stride Length & Step Width \\
Keypoints & Method &  &  &  &  &  &  &  \\
\midrule
\multirow[c]{12}{*}{MMPoseHalpe} & Explicit Optimization & 82.4 & 38.4 & 85.3 & 39.9 & 51.8 & 35.3 & 13.4 \\
 & Explicit Optimization KP Conf, MaxHuber=10 & 23.5 & \bfseries 9.9 & 19.6 & \bfseries 14.4 & 18.1 & 19.4 & 9.6 \\
 & Implicit Optimization & 19.1 & 10.5 & \bfseries 16.5 & 14.9 & 17.3 & 15.7 & \bfseries 9.4 \\
 & Implicit Optimization $\gamma=0.3$ & 108.3 & 46.5 & 112.1 & 47.2 & 52.1 & 42.9 & 25.0 \\
 & Implicit Optimization KP Conf & 28.1 & 10.7 & 24.5 & 15.2 & 33.1 & 42.5 & 13.5 \\
 & Implicit Optimization KP Conf, MaxHuber=10 & 21.7 & 10.3 & 20.6 & 14.7 & 18.5 & 15.4 & 10.5 \\
 & Implicit Optimization, MaxHuber=10 & 32.6 & 13.4 & 18.6 & 14.5 & 38.3 & \bfseries 15.1 & 13.7 \\
 & Robust Triangulation & 24.7 & 11.0 & 22.5 & 16.4 & 18.3 & 16.8 & 10.0 \\
 & Robust Triangulation $\gamma=0.3$ & 27.0 & 11.2 & 26.1 & 16.0 & 18.7 & 19.8 & 10.2 \\
 & Robust Triangulation $\sigma=100$ & 24.3 & 11.1 & 22.0 & 16.4 & 17.5 & 16.2 & 9.7 \\
 & Robust Triangulation $\sigma=50$ & 29.7 & 16.9 & 24.0 & 15.7 & \bfseries 17.1 & 17.0 & 9.9 \\
 & Triangulation & 36.3 & 16.9 & 38.1 & 22.5 & 37.4 & 41.9 & 13.6 \\
\midrule
\multirow[c]{12}{*}{OpenPose HR} & Explicit Optimization & 46.8 & 27.6 & 57.6 & 37.1 & 30.2 & 41.1 & 18.5 \\
 & Explicit Optimization KP Conf, MaxHuber=10 & 257.2 & 147.3 & 224.2 & 89.1 & 532.1 & 299.9 & 360.4 \\
 & Implicit Optimization & 29.9 & 14.1 & 41.0 & 21.1 & 39.8 & 32.3 & 16.0 \\
 & Implicit Optimization $\gamma=0.3$ & 68.9 & 40.9 & 80.2 & 47.1 & 27.1 & 25.7 & 22.7 \\
 & Implicit Optimization KP Conf & 37.1 & 15.8 & 48.7 & 23.9 & 49.8 & 32.8 & 15.5 \\
 & Implicit Optimization KP Conf, MaxHuber=10 & 47.6 & 17.9 & 54.9 & 25.0 & 73.1 & 41.0 & 20.9 \\
 & Implicit Optimization, MaxHuber=10 & 41.8 & 17.5 & 46.7 & 22.3 & 65.2 & 40.7 & 22.8 \\
 & Robust Triangulation & 57.1 & 19.6 & 91.4 & 31.8 & 23.1 & 138.7 & 24.7 \\
 & Robust Triangulation $\gamma=0.3$ & 30.0 & 15.0 & 35.5 & 21.5 & 21.4 & 17.0 & 11.8 \\
 & Robust Triangulation $\sigma=100$ & 57.0 & 20.1 & 91.2 & 32.7 & 23.1 & 131.9 & 15.6 \\
 & Robust Triangulation $\sigma=50$ & 57.8 & 24.8 & 92.2 & 36.0 & 25.1 & 132.3 & 15.7 \\
 & Triangulation & 56.1 & 27.1 & 56.4 & 31.9 & 49.3 & 80.6 & 23.3 \\
\midrule
\multirow[c]{12}{*}{OpenPose LR} & Explicit Optimization & 79.4 & 51.6 & 85.9 & 45.5 & 35.7 & 56.1 & 22.9 \\
 & Explicit Optimization KP Conf, MaxHuber=10 & 164.0 & 86.6 & 177.1 & 74.2 & 365.5 & 195.2 & 211.7 \\
 & Implicit Optimization & 31.8 & 15.0 & 41.8 & 22.0 & 36.6 & 30.6 & 15.2 \\
 & Implicit Optimization $\gamma=0.3$ & 88.0 & 59.4 & 92.3 & 61.0 & 32.0 & 51.5 & 21.9 \\
 & Implicit Optimization KP Conf & 35.1 & 15.7 & 46.2 & 24.5 & 43.1 & 32.2 & 14.9 \\
 & Implicit Optimization KP Conf, MaxHuber=10 & 44.0 & 18.3 & 54.0 & 25.5 & 70.4 & 37.2 & 20.3 \\
 & Implicit Optimization, MaxHuber=10 & 41.0 & 17.9 & 49.5 & 22.7 & 62.8 & 37.8 & 22.6 \\
 & Robust Triangulation & 32.0 & 16.1 & 62.9 & 32.4 & 21.4 & 99.6 & 13.2 \\
 & Robust Triangulation $\gamma=0.3$ & 29.2 & 15.9 & 37.1 & 24.9 & 21.7 & 18.1 & 13.0 \\
 & Robust Triangulation $\sigma=100$ & 30.3 & 17.6 & 61.7 & 45.6 & 21.1 & 99.6 & 13.1 \\
 & Robust Triangulation $\sigma=50$ & 34.7 & 19.7 & 65.0 & 37.2 & 21.1 & 99.8 & 13.7 \\
 & Triangulation & 45.3 & 24.7 & 46.7 & 28.9 & 43.6 & 66.9 & 22.0 \\
\bottomrule
\end{tabular}

\end{table*}

\end{document}